\documentclass[letterpaper, 10 pt, conference]{ieeeconf}
\IEEEoverridecommandlockouts  
\usepackage{graphicx} 
\usepackage{amsmath}
\usepackage{comment}

\newtheorem{theorem}{\textbf{Theorem}}[section]
\newtheorem{remark}[theorem]{\textbf{Remark}}
\newtheorem{assumption}[theorem]{\textbf{Assumption}}

\usepackage{amssymb}
\usepackage{bm}
\usepackage{longtable}
\usepackage{subcaption}
\usepackage{balance}

\title{\LARGE \bf Agree to Disagree: Consensus-Free Flocking under Constraints}

\author{Peter Travis Jardine$^{1}$, and Sidney Givigi$^{1}$
\thanks{$^{1}$  P.\ Jardine and S.\ Givigi are with the School of Computing, 
        Queen's University, Kingston, ON K7L 3N6 Canada 
        {\tt\small p.jardine@queensu.ca,  sidney.givigi@queensu.ca}}%
}

\begin{document}

\maketitle

\begin{abstract}
Robots sometimes have to work together with a mixture of partially-aligned or conflicting goals. Flocking — coordinated motion through cohesion, alignment, and separation — traditionally assumes uniform desired inter-agent distances. Many practical applications demand greater flexibility, as the diversity of types and configurations grows with the popularity of multi-agent systems in society. Moreover, agents often operate without guarantees of trust or secure communication. Motivated by these challenges we update well‐established frameworks by relaxing this assumption of shared inter-agent distances and constraints. Through a new form of constrained collective potential function, we introduce a solution that permits negotiation of these parameters. In the spirit of the traditional flocking control canon, this negotiation is achieved purely through local observations and does not require any global information or inter-agent communication. The approach is robust to semi-trust scenarios, where neighbouring agents pursue conflicting goals. We validate the effectiveness of the approach through a series of simulations.
\end{abstract}

\section{Introduction}

Multi-agent robotics research is motivated by the simple idea that tasks can be completed better, faster, or with less effort when multiple agents work together. Swarm robotics embodies this principle through decentralized, asynchronous systems of simple, quasi-identical agents operating under local rules \cite{Beni-2005}. Their collective behavior is emergent, with properties not present in the individuals. 

An example of swarm behavior is \textit{flocking}, initially formulated in \cite{reynolds1987} and later integrated into control theory in \cite{olfati2006}. Flocking 
involves group objectives (e.g., target tracking or maintenance of structural qualities) and typically results in agents forming an evenly-spaced \textit{lattice} structure. The practical applications of flocking strategies are numerous, ranging from mobile communications, remote monitoring and data collection, distributed resource delivery, search and rescue, and military operations \cite{10373773}. A recent review suggested a need for more comprehensive standardization of flocking methodologies to enable greater adoption in real-world applications \cite{NEDJAH2019100565}. This paper examines scenarios where agents pursue both shared and individual goals with limited trust \cite{10555633,akbari2024factorgraphmodeltrust}.

\subsection{Related Work}

As the field draws from a broad range of related domains, 
it is rich with diverse methodologies, often achieving similar outcomes in different ways. Due to the work of \cite{olfati2004consensus, olfati2003consensus, Wen2012}, gradient- and consensus-based methods have drawn considerable attention in flocking research. In such approaches, interactions between agents are typically represented as a graph, while performance and stability are analyzed in terms of control-Lyapunov functions. Recent work has applied flocking techniques to produce homeomorphic curves~\cite{Jardine2025}. While significant work has investigated the case of time-varying topologies \cite{1431045}, a common assumption is that the graph is undirected and connected at all times \cite{zhang2011general}. This early work generally relied on two assumptions: agents share common navigation objectives and separation parameters. Later work relaxed the former condition using \textit{pinning} control, demonstrating only a subset of the agents need be informed of navigation goals to remain stable \cite{Su-2009, Gao-2017}. 
Work in flocking has focused on finding control solutions such that \cite{olfati2006}:
\begin{equation} \label{eq:sep0}
    \|\mathbf{x}_i(t) - \mathbf{x}_j(t)\| = d~\forall~j~\in~\mathcal{V}_i,
\end{equation}
\noindent where $\mathbf{x}_i(t)$ and $\mathbf{x}_j(t)$ are the positions of agents $i$ and $j$ at time $t$, respectively; $\mathcal{V}_i$ defines the neighbourhood of agent $i$; and $d$ is the desired separation. Such work has typically assumed $d$ is fixed for all agents (either explicitly or indirectly through prescribed interaction ranges) or focused on fault-tolerant methodologies that identify conflicting goals \cite{DONG2015197,8854221}. Recent work has extended into strategies which attempt to counter malicious agents by modeling their parameters and adapting the structure of the entire flock \cite{10264142}; such methods prioritize connectivity maintenance by allowing a single malicious agent to drive the trajectory of the entire flock.
We  build on this work to consider the specific case when the initial desired separation between agents is not necessarily equal for all agents in the flock and must be negotiated over time, or: 
\begin{equation} \label{eq:stab_dij}
\|\mathbf{x}_i(t) - \mathbf{x}_j(t)\| = d_{ij}(t)~\forall j\in\mathcal{V}_i~|
~{d}_{ij}(0) \not\equiv {d}_{ji}(0).
\end{equation}

Rather than focusing on purely malicious agents, we consider semi-trust scenarios, where agents cooperate within their respective design parameters but do not implicitly trust one-another or communicate directly \cite{10555633, akbari2024factorgraphmodeltrust}. We address these challenges through the construction of a novel constrained collective potential function that negotiates separation parameters while relying on local observations of neighbouring agents. This requires that each agent model the dynamics of their neighbour from local observations. Agents must also protect themselves from the influence of neighbours that operate outside their individual constraints. This generalization affords greater robustness to agents with different or time-varying parameters, physical constraints, or external influences. Examples include groups with dissimilar physical dynamics (such as a mix of ground, aerial, and waterborne robots) that must cooperate or maintain proximity in order to preserve inter-agent communications channels. Even similar robots may have different quality or types of hardware that require different operating ranges. Considering such heterogeneity in agent configuration is important as the scale of commercial robotics application grows. If not properly managed, such irregularities have been shown to lead to destabilization or collapse of the flock \cite{tanner2003stable}. 

\subsection{Agent Dynamics} \label{sec:dyn}

Let us consider agent $i \in \mathcal{V}$ at time $t$ with position $\mathbf{x}_i(t)~\in~\mathbb{R}^2$, velocity $\mathbf{v}_i(t)~\in~\mathbb{R}^2$, and inputs $\mathbf{u}_i(t)~\in~\mathbb{R}^2$ governed by the following \textit{double integrator} dynamics:
\begin{equation} \label{eq:dyn}
\begin{matrix}
\dot{\mathbf{x}}_i(t) = \mathbf{v}_i(t), \\
\dot{\mathbf{v}}_i(t) = \mathbf{u}_i(t),
\end{matrix}
\end{equation}
\noindent where $\mathbf{u}_i(t)=\ddot{\mathbf{x}}_i(t)$ is a vector of accelerations. Such dynamics are applicable to a wide range of dynamic systems.


\subsection{Connectivity} \label{sec:con}

Consider a set of agents as an undirected connected graph $\mathcal{G}=(\mathcal{V}, \mathcal{E})$ with vertices $\mathcal{V}$ and edges $\mathcal{E} \subseteq \{(i,j): i,j \in \mathcal{V}, j \neq i\}$. An edge exists between agents $i$ and $j$ if they are in the neighbourhood of each other. Specifically, the neighbourhood of agent $i$ is defined as
\begin{equation} \label{eq:connectivity}
    \mathcal{V}_i = \{ j \in \mathcal{V} : \| \mathbf{x}_i(t) - \mathbf{x}_j(t)\| < {r}_{i}\},
\end{equation}
\noindent where ${r}_{i}$ is the fixed range at which agent $i$ can sense or communicate with neighbouring agents. Note the necessary relationship between this range and the desired separation -- agents must be able to sense or communicate with neighbours to pursue equilibrium on separation.  
\textit{Adjacency matrix} $A$ describes the connections between vertices, where matrix element $A_{ij}$ is $1$ if $i$ and $j$ are neighbours or $0$ otherwise:
\begin{equation}
A_{ij} = 
    \begin{cases} 
        1 & \text{if \(\exists\) edge between } i \text{ and } j, \\ 
        0 & \text{otherwise}. 
    \end{cases}
\end{equation}

A \textit{component} set of the graph is a set with no neighbours outside itself. Graph components can be considered subgraphs, each of which is not part of any other subgraph. When using pinning control, where certain agents (vertices) are selected to drive the motion of the larger network (graph), the concept of a component becomes very important, as each component requires at least one pin in order to drive its constituent vertices towards a target. Pin vertices are typically selected based on measures of centrality.

There are numerous measures to describe the centrality of specific vertices within a graph. Examples include degree, controllability, betweenness, and eigenvalue centrality. We select matrix-form \textit{degree centrality} as it reflects the relative importance of a vertex based on its connectivity with other vertices and is easily computed from $A$ as follows \cite{Nozari-2019, Pasqualetti-2014}:
\begin{equation} \label{eq:degcentrality}
\Delta_{i} = \sum_{j \in \mathcal{V}} A_{ij},
\end{equation}
\noindent where $\Delta_{i}$ is the degree centrality for vertex $i$ and:
\begin{equation}
    \Delta
    =
    \begin{bmatrix}
        \Delta_1 & \dots & 0 \\
        \vdots & \ddots & 0 \\
        0 & 0 & \Delta_n
    \end{bmatrix}.
\end{equation}


\section{Homogeneous Lattice Flocking} \label{sec:lattice}

This section draws on previous work to formulate homogeneous lattice flocking. Its basic elements (cohesion, alignment, and navigation) are presented in their general form as energy potential functions. This formulation serves as a basis on which we expand into heterogeneous lattices in Section \ref{sec:hetlatscale}. For convenience, we temporarily remove the time notation $(t$) throughout this section.

\subsection{Cohesion} \label{sec: Cohesion}

As described above, there exist numerous solutions in the literature to achieve cohesion in the form of a lattice, as defined by \eqref{eq:sep0}.  Examples include variations of Lennard-Jones \cite{Wen2012, Moscato2024}, Olfati-Saber 
\cite{olfati2006}, Morse \cite{Morse1929}, Soft Repulsion \cite{Rogers2014}, Mixed Exponential and Polynomial \cite{Schweighofer2008}, Riesz \cite{Landkof1972}, Gaussian \cite{Franz2007},  reinforcement learning~\cite{Liu2024}, elliptical lattices~\cite{Wang2023}, and Gromacs Soft-Core \cite{Gapsys2012} approaches. In all cases, cohesion can be achieved by each agent $i$ minimizing a specially-constructed energy potential function ${V}^c_i(\mathbf{z}^c_i)$ using gradient descent. We represent the control solution as \cite{Wen2012}:
\begin{equation} \label{eq:gradcontrol}
    \dot{\mathbf{x}}_i = -\nabla_{\mathbf{x}_i}{V}^c_i(\mathbf{z}^c_i),
\end{equation}
\noindent where $\mathbf{z}^c_i$ is a tuple containing states specific to the design of ${V}^c_i(\mathbf{z}^c_i)$, i.e., $\mathbf{z}^c_i = (\mathbf{x}_i , \mathbf{x}_j)$. Given $\mathbf{z}^c_i$, desired separation $d_{ij}$, and tunable parameter $k_c>0$, we consider a modified version of the commonly-used Lennard-Jones potential \cite{Moscato2024}:
\begin{equation} \label{eq:lj}
{V}^c_{i}(\mathbf{x}_i , \mathbf{x}_j) = \left( \frac{k_cd_{ij}^6}{2\| \mathbf{x}_i - \mathbf{x}_j\|} \right)^{12} - \left( \frac{k_c}{\| \mathbf{x}_i - \mathbf{x}_j\|} \right)^6 ,
\end{equation}
\noindent for which we may define the gradient as follows:
\begin{equation}
\nabla_{\mathbf{x}_i}{V}^c_i( \mathbf{x}_i , \mathbf{x}_j) = \left(\frac{-6k_cd_{ij}^6}{\mathbf{\| \mathbf{x}_i - \mathbf{x}_j\|}^{13}} + \frac{6k_c}{\mathbf{\| \mathbf{x}_i - \mathbf{x}_j\|}^7} \right) \mathbf{\hat{x}}_{ij},
\end{equation}
\noindent with
\begin{equation*}
    \mathbf{\hat{x}}_{ij}=\left(\frac{\mathbf{x}_i - \mathbf{x}_j}{\mathbf{\mathbf{\| \mathbf{x}_i - \mathbf{x}_j\|}}}\right),
\end{equation*}and design a stabilizing control solution of the form \eqref{eq:gradcontrol}
that converges on equilibrium point where $\| \mathbf{x}_i - \mathbf{x}_j\|=d_{ij}$. 
Note that, in the above formulation, $d_{ij}$ is assumed to be a fixed parameter. In later sections we will relax this condition, which requires a modified gradient function that accounts for a time-varying $d_{ij}(t)$.

\subsection{Alignment} \label{sec: Align}

Similar to the problem of cohesion described above, there exist numerous approaches to achieve alignment among agents. Broadly, alignment is the tendency to adopt the average directionality (i.e., heading) of neighbours. Aligning with the notation used above, let us encapsulate the evolution of the states using the following gradient-based control policy:
\begin{equation} \label{eq:gradcontrol_align}
    \dot{\mathbf{v}}_i = -\nabla_{\mathbf{v}_i}{V}^a_i(\mathbf{z}^a_i)
\end{equation}
\noindent where $\mathbf{z}^a_i$ is a tuple containing states specific to the design of ${V}^a_i(\mathbf{z}^a_i)$, i.e., $\mathbf{z}^a_i=( \mathbf{x}_i , \mathbf{x}_j , \mathbf{v}_i , \mathbf{v}_j)$. Drawing on previous work~\cite{1272912}, let us define
\begin{equation} \label{eq:Va}
    V^a_i( \mathbf{z}^a_i) = \frac{1}{2} k_a\sum_{j \in \mathcal{V}_i} \rho_\beta(\| \mathbf{x}_i - \mathbf{x}_j\|_\epsilon) \|\mathbf{v}_i - \mathbf{v}_j\|^2,
\end{equation}
\noindent where $k_a>0$ is a tunable parameter, $\|\mathbf{x}\|_\epsilon = \frac{\epsilon\| \mathbf{x}_j - \mathbf{x}_i\|}{\sqrt{1 + r^2}-1}$ is a norm that considers the sensing radius $r$ of the agent scaled by parameter $\epsilon > 0$ and:
\begin{equation} \label{eq:align1}
 \rho_\beta(\alpha) = 
    \begin{cases}
        1 & 0 \leq \alpha < \beta,\\
        \frac{1}{2}( 1 + \cos(\pi\frac{\alpha-\beta}{1-\beta})) & \beta < \alpha \leq 1, \\
        0 & \text{otherwise},
    \end{cases}
\end{equation}
\noindent is a smooth function that varies between $0$ and $1$, depending on the relative positions of the agents. $\beta\in[0,1]$ is a parameter chosen by the designer. 
The gradient then is
\begin{equation}
 \nabla_{\mathbf{v}_i} V^a_i( \mathbf{z}^a_i) = 
 k_a\sum_{j \in \mathcal{V}_i} \rho(\|\mathbf{x}_i - \mathbf{x}_j\|_\epsilon) (\mathbf{v}_i - \mathbf{v}_j).
\end{equation}

\subsection{Navigation} \label{Sec: Pinning}

A central finding in \cite{olfati2006} is the need to account for group objectives in flocking to establish and maintain global cohesion, especially when the flock is not initially connected. Without a shared objective, lattice formation may lead to fragmentation. Previous work has typically addressed this by defining a group objective in the form of a shared navigation term as follows:
\begin{equation} \label{eq:gradcontrol_nav}
    \dot{\mathbf{x}}_i = -\nabla_{\mathbf{x}_i}{V}^n_i(\mathbf{z}^n_i)
     -\nabla_{\mathbf{v}_i}{V}^n_i(\mathbf{z}^n_i),
\end{equation}
\noindent where $\mathbf{z}^n_i$ is a tuple containing states specific to the design of ${V}^n_i(\mathbf{z}^c_i)$, nominally at least positions $\mathbf{x}_i$ and velocities $\mathbf{v}_i$, as well as references $\mathbf{x}_r$ and $\mathbf{v}_r$. We adopt this design philosophy, but reduce the number of agents that must perform this task by incorporating pinning control as follows:
\begin{equation} \label{eq:Vn}
    V^n_i(\mathbf{x}_i, \mathbf{v}_i) = \frac{1}{2}\kappa(i)(k_{{n}_{x}}\|\mathbf{x}_i - \mathbf{x}_r\|^2 + k_{{n}_{v}}\|\mathbf{v}_i - \mathbf{v}_r\|^2),
\end{equation}
\noindent where $\mathbf{x}_r$ and $\mathbf{v}_r$ are fixed reference states and velocities, $k_{n_x}$ and $k_{n_v}$ are tunable parameters, and 
\begin{equation}
\kappa(i) =
    \begin{cases}
         1 & \text{if}~i~\in~\mathcal{P}, \\
         0 & \text{otherwise},
    \end{cases}
\end{equation}
\noindent where the pin set $\mathcal{P}$ is defined below. Control of complex networks often benefits from specialization of tasks for different agents. A common such approach is to pin the dynamics of agents to specific states using pinning control \cite{9226139}, so that certain agents drive the behavior of the larger network (e.g., reference tracking), while the remaining agents focus on local interactions (e.g., alignment or cohesion). 

In our approach, pins are selected in response to the evolving structure of the graph (specifically, the shifting influence of agents in individual graph components) and guide the agents towards convergence into a single structure. At the same time, the lattice structure is formed and maintained between these agents using the methods described earlier. 
Recalling graph $\mathcal{G} = (\mathcal{V},\mathcal{E})$ from Section \ref{sec:con}, let us denote components of $\mathcal{G}$ as:
\begin{equation}
    \{ \mathcal{C}_1, \mathcal{C}_2, \ldots, \mathcal{C}_m \} \subseteq \mathcal{V}, 
\end{equation}
\noindent where $m$ is the total number of components. The set of all vertices is the union of these components and $\mathcal{C}_g\bigcap\mathcal{C}_h=\varnothing$ for $g \neq h$. 
Recalling \eqref{eq:degcentrality}, for each component we select  vertex $i_k^* \in \mathcal{C}_k \subset \mathcal{V}$ with maximum degree centrality: 
\begin{equation}
i_k^* = \arg\max_{i \in \mathcal{C}_k} \Delta_{i}.
\end{equation}

We describe the union of these selected vertices as the set:
\begin{equation}
\mathcal{P}_d = \{i_1^*, i_2^*, \ldots, i_m^*\},
\end{equation}
\noindent where, for our application, $\mathcal{P}_d$ contains one pin vertex from each component (i.e., the agent with maximum degree centrality). As investigated in \cite{leaf-2021}, pinning \textit{leaf} nodes (i.e., those with degree 1) can provide greater coupling between unpinned nodes and improve overall convergence properties with comparably minimal perturbations on the greater network dynamics. We define the set of all leafs as:
\begin{equation}
\mathcal{P}_{l} = \bigcup_{k=1}^m \mathcal{P}_{l,k}\quad, \quad
\mathcal{P}_{l,k} = \{i \in \mathcal{C}_k \mid \Delta_i = 1\}. 
\end{equation}

Finally, we define the set of all pins as
$
\mathcal{P} = \mathcal{P}_{d} \cup \mathcal{P}_{l}
$
and compute the gradient as:
\begin{equation} \label{eq:dVn}
    \nabla V^n_i(\mathbf{x}_i, \mathbf{v}_i) = \kappa(i)(k_{n,x}(\mathbf{x}_i - \mathbf{x}_r) + k_{n,v}(\mathbf{v}_i - \mathbf{v}_r)).
\end{equation}

\subsection{Summary}

The conditions under which the combined implementations of Sections \ref{sec: Cohesion}, Section \ref{sec: Align}, and Section \ref{Sec: Pinning} are stable -- assuming shared separation parameters -- are well-established in the literature \cite{olfati2006} \cite{Moscato2024}. Without loss of generality, for the case when $d_{ij} = d~\forall(i,j)\in\mathcal{V}$ (i.e., homogeneous flocking), 
let us
define a stable cohesion, alignment, and navigation potential function as follows:
\begin{eqnarray} \label{eq:Vfi}
V^f_i(\mathbf{x}_i , \mathbf{x}_j , \mathbf{v}_i , \mathbf{v}_j, d) 
& = &
\sum_{j\in \mathcal{V}_i}
\Big(
{V}^c_{i}(\mathbf{x}_i , \mathbf{x}_j)
\nonumber\\ & + &
V^a_i(\mathbf{x}_i , \mathbf{x}_j , \mathbf{v}_i , \mathbf{v}_j)
\nonumber\\ & + &
V^n_i(\mathbf{x}_i, \mathbf{v}_i) \Big),
\end{eqnarray}
\noindent for which the stabilizing controllers are defined above. 

\begin{assumption}[Stable Homogeneous Lattice Flocking] \label{assum:Vtilde1}
Given agents governed by the dynamics at \eqref{eq:dyn}, connected as defined in \eqref{eq:connectivity}, and homogeneous lattice parameters (i.e., $d_{ij} = d~\forall(i,j)\in\mathcal{V}$), there exists a continuously differentiable potential function ${V}^f_i({\mathbf{z}^f_i})$ and standard gradient descent control law \(
{\dot{\mathbf{z}}^f_i} = -\nabla_{\mathbf{z}^f_i} {V}^f_i(\mathbf{z}^f_i)
\) for all agents $i \in \mathcal{V}$, such that for suitably constructed Lyapunov candidate function ${V}^f_{L}(t)=\sum_{i\in\mathcal{V}} V_L^i(\mathbf{z}^f_i)$ such that:
\begin{equation} \label{eq:VfL_stab}
\begin{matrix} 
{V}^i_{L}(\mathbf{z}_i^{f*}) & = 0 , \\
{V}^i_{L}(\mathbf{z}^f_i) & > 0~\forall~\mathbf{z}_i \neq \mathbf{z}_i^{f*} &, \\
\dot{{V}}^i_{L}(\mathbf{z}^f_i) & < 0~\forall~\mathbf{z}_i \neq \mathbf{z}_i^{f*}&,
\end{matrix}
\end{equation}
\noindent where $\mathbf{z}_i^{f*}$ is an equilibrium point satisfying \eqref{eq:sep0}. That is to say, there exists a gradient-based control law that stabilizes each one of the agents and, therefore, the system around an equilibrium configuration where all agents are separated by desired separation $d$.  
\end{assumption} 

\begin{remark}
Assumption \ref{assum:Vtilde1} is intentionally broad, to capture the various types of lattice-forming flocking methodologies available. However, in the applications that follow, we rely on the specific implementations presented in the preceding sections. 
\end{remark}

\begin{figure}
    \centering
    \begin{subfigure}[t]{0.45\columnwidth}
        \centering
        \includegraphics[width=\textwidth]{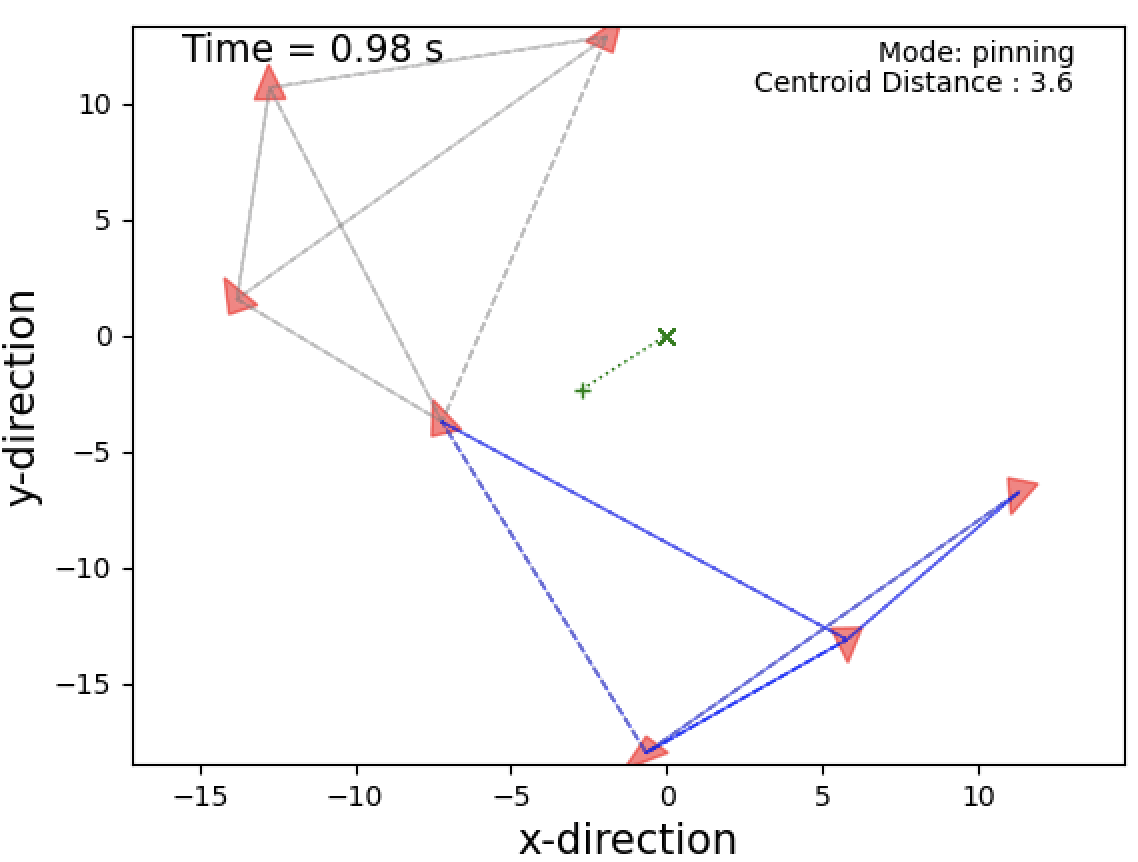}
        \caption{Initial state of flock.}
        \label{fig:break1}
    \end{subfigure}
    \begin{subfigure}[t]{0.45\columnwidth}
        \centering
        \includegraphics[width=\textwidth]{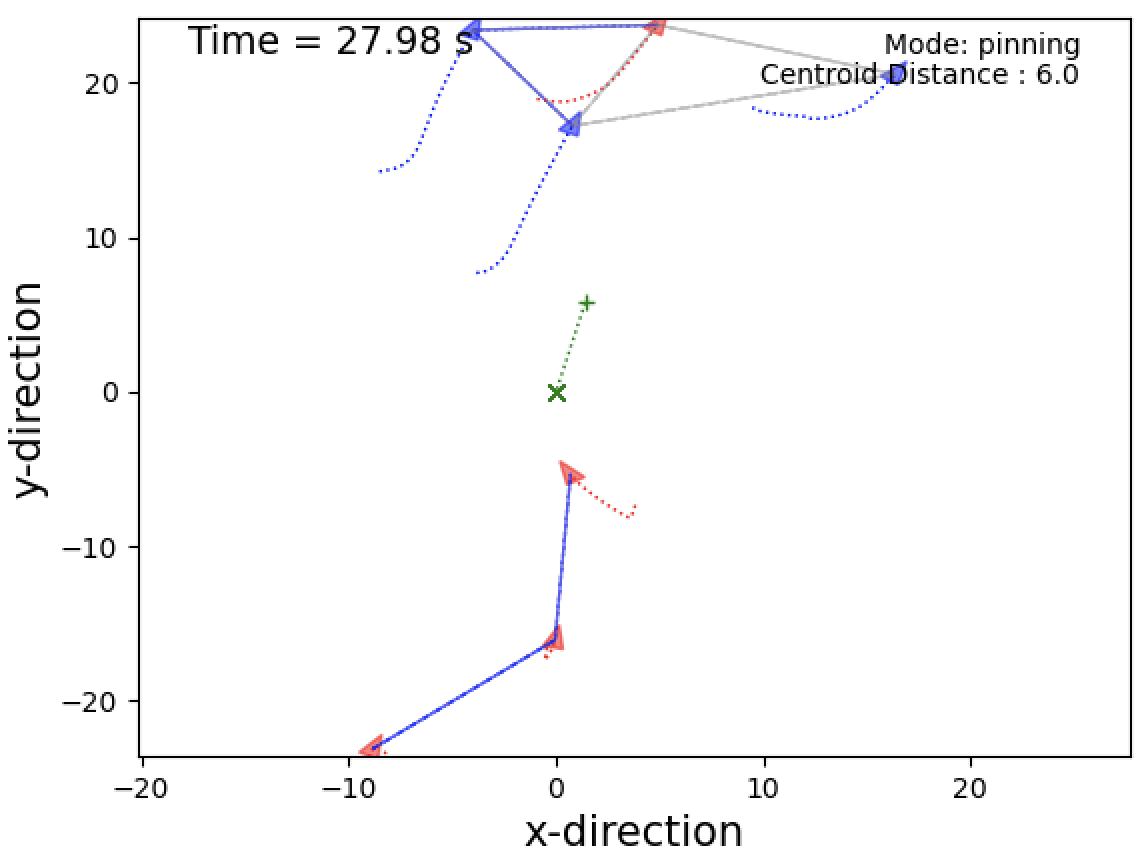}
        \caption{Flock collapses.}
        \label{fig:break2}
    \end{subfigure}
    \caption{Agents unable to reach equilibrium when configured with different separation distances ($d_{ij}$).}
    \label{fig:break}
\end{figure}

We now relax the condition on lattice parameter homogeneity and observe the effect this has on the flock. Fig. \ref{fig:break} demonstrates the effect of this relaxation for $7$ agents. Each agent was configured to form a lattice using \eqref{eq:lj}, but with fixed, randomly-distributed values of $5<d_{ij}<13$. Note that the agents diverge from one-another and the number of total connections decreases. While this undesirable behaviour is similar to the fragmentation reported in \cite{olfati2006}, it is caused by an entirely different phenomenon. Whereas fragmentation occurs due to lack of a shared navigation term, this instability occurs due to different lattice parameters (i.e., different values of $d_{ij}$). Each agent is attempting unsuccessfully to converge on different equilibrium points.
In practice, agents need to negotiate compatible equilibria locally to achieve global stability of the flock. We consider this to be a distributed optimization problem where each agent must select its own $d_{ij}(t)$ equilibrium point(s) while also carrying out the control necessary to stabilize its position, noting this may result in different separations between neighbours. We further assume the agents are unable to share these parameters through communication and must instead infer them from local observations. 

\section{Heterogeneous Lattice Flocking} \label{sec:hetlatscale}

The discussion above describes the breadth of lattice formation strategies available and the conditions under which they are stable. A common assumption of these approaches is a fixed and shared lattice scale parameter $d_{ij}$. Let us consider a time-varying desired separation of agent $i$ from agent $j$ as $d_{ij}(t)$, noting that $d_{ij}(0) \not\equiv d_{ji}(0)$ initially. We derive a new parameter $\hat{d}_{ji}(t)$ from the following filter based on the observed distance between agents $i$ and $j$:
\begin{equation} \label{filter}
    \dot{\hat{d}}_{ji}(t) = \frac{1}{\tau}\left( \|\mathbf{x}_i(t) - \mathbf{x}_j(t)\| -\hat{d}_{ji}(t) \right), 
\end{equation}
\noindent with $\hat{d}_{ji}(0) = {d}_{ij}(0)$. $\tau > 0$ is the time constant for the filter and $\hat{d}_{ji}$ represents an estimate of the neighbouring agent's desired separation. The purpose of \eqref{filter} is for each agent to model the dynamics of its neighbour's parameters (hence the swapped subscripts of $\hat{d}_{ji}$) based on local observations, thereby removing the necessity to establish trusted, secure communications. 

For the purpose of achieving equilibrium with neighbouring agents, each agent is able to adjust its desired separation within the following bounds $d_{i,\min}<d_{ij}(t)<d_{i,\max}$. Note that each agent may have different bounds. Our goal is to design a strategy that achieves this equilibrium without violating these constraints. For convenience, we remove the notation for time ($t$) throughout the remainder of this section.
\subsection{Custom bump function}
\begin{figure}
    \centering
    \includegraphics[width=0.9\columnwidth]{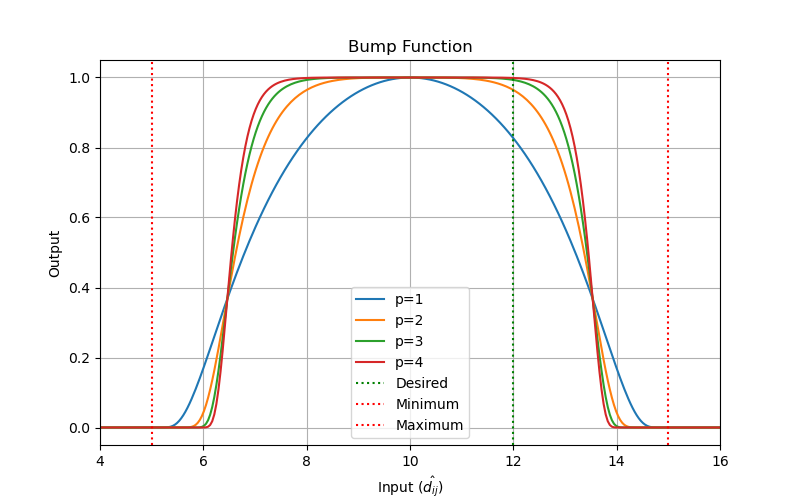} 
    \caption{Bump function with $d_{i,\min}=5$ and $d_{i,\max}=15$.}
    \label{fig:bump}
\end{figure}

Let us define a new parameter:
\begin{equation}
    \lambda = \frac{2 \left(\hat{d}_{ji} - \frac{d_{i,\min} + d_{i,\max}}{2}\right)}{d_{i,\max} - d_{i,\min}},
\end{equation}
\noindent and design the following custom smooth \textit{bump} function:
\begin{equation} \label{eq:bump}
V_i^b(\hat{d}_{ji}) =
    \begin{cases}
        e^{\left(-\frac{\lambda^2}{1 - \lambda^2}\right)^p}, & \text{if } d_{i,\min} < \hat{d}_{ji} < d_{i,\max}, \\
        0, & \text{otherwise},
    \end{cases}
\end{equation}
\noindent where $p>1$ adjusts the flatness of the bump function at its peak and $d_{i,\min}<\hat{d}_{ji}<d_{i,\max}$ are constraints around the desired separation. Fig.~\ref{fig:bump} illustrates the behavior of the bump function at various values of $p$, noting that the function is smooth and greater than $0$ when $\hat{d}_{ji}$ is within bounds. 
The gradient of \eqref{eq:bump} 
$d_{i,\min} < \hat{d}_{ji} < d_{i,\max}$ is 
{
\begin{equation}\label{eq:bumpgrad}
\nabla_{\hat{d}_{ji}} V^b_i(\hat{d}_{ji}) =
V_i^b(\hat{d}_{ji}) \left(-\frac{4\lambda \cdot p \cdot \left(\frac{\lambda^2}{1 - \lambda^2}\right)^{p-1}}{(1 - \lambda^2)^2 (d_{i,\max} - d_{i,\min})}\right),
\end{equation}
}
\begin{figure}
    \centering
    \includegraphics[width=0.9\columnwidth]{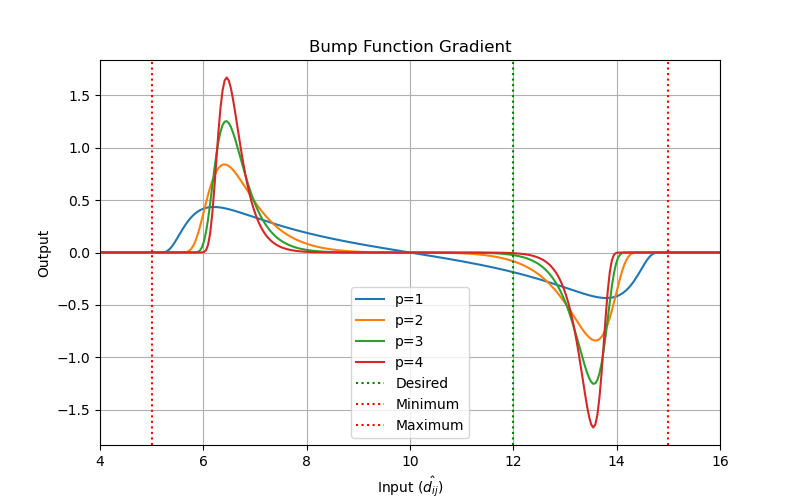} 
    \caption{Gradient of bump function with $d_{i,\min}=5$ and $d_{i,\max}=15$.}
    \label{fig:bumpd}
\end{figure}
Note that \eqref{eq:bumpgrad} depicted in Fig.~\ref{fig:bumpd} is greatest approaching the bounds, prior to decreasing back to zero. 
\subsection{Constrained Collective Potential Function} \label{sec:controlpols}

This section presents a new constrained collective potential function, which negotiates inter-agent distance values without violating individual agent constraints. Let us define a controller of the form:
\begin{equation} \label{eq:d_c}
    \dot{{d}}_{ij} = -\nabla_{{d}_{ij}}{V}^d_{ij}(\mathbf{z}^d_{ij}),
\end{equation}
\noindent where ${\mathbf{z}}^d_{ij}$ is a tuple containing $d_{ij}$, and $\hat{d}_{ji}$. We define the following potential function, which contains the custom bump function \eqref{eq:bump} above: 
\begin{equation} \label{eq:Vd_PF}
    {V}^d_{ij}(d_{ij}, \hat{d}_{ji}) = \frac{1}{2}k_d V^b_i(\hat{d}_{ji})({d}_{ij}-\hat{d}_{ji})^2,
\end{equation}
\noindent where $k_d>0$ is a tunable parameter. Note that whereas previous potential functions have been defined for agent $i$, \eqref{eq:Vd_PF} is defined per agent $i$-$j$ pair. This potential function increases with the error between variables $\hat{d}_{ji}$ and ${d}_{ij}$ when $\hat{d}_{ji}$ is within bounds. Importantly, the potential is zero outside these bounds. We compute the gradient as follows:
\begin{equation} \label{eq:dij_feedback}
    \nabla_{{d}_{ij}} {V}^d_{ij}(d_{ij}, \hat{d}_{ji}) = 
    k_d V^b_i(\hat{d}_{ji}) ({d}_{ij}-\hat{d}_{ji}).
\end{equation}

\subsection{Summary}


The formulation above builds on well-established gradient-based flocking methods with known convergence properties for the homogeneous case. To accommodate heterogeneity, we permit local negotiation of separation parameters within explicitly defined bounded regions, while otherwise preserving the structure of these dynamics. 
Drawing on \eqref{eq:Vfi} and Assumption~\ref{assum:Vtilde1} for fixed parameter $d_{ij}(t)=d~\forall~(i,j)\in\mathcal{V}$, our approach induces a shift in the equilibrium points relative to this notional homogeneous case. This shift reflects locally negotiated inter-agent equilibria of inter-agent distances rather than a globally shared fixed parameter.

\section{Simulation Results and Discussions}\label{sec:Results}
The flocking techniques described above were implemented in several simulations. The first simulation demonstrates effective assembly of a large homogeneous flock from random initial conditions. In the subsequent simulations, each agent pursues a mix of shared and distinct goals. Namely, the agents share a common navigation goal but are initiated with conflicting lattice separation parameters and bounds. At no time are the agents able to communicate their parameters; instead, each develops a unique model of its neighbours dynamics in order to negotiate an equilibrium. 

\subsection{Homogeneous Assembly}
\begin{figure}[!ht]
    \centering
    \begin{subfigure}[t]{0.45\columnwidth}
        \centering
        \includegraphics[width=\textwidth]{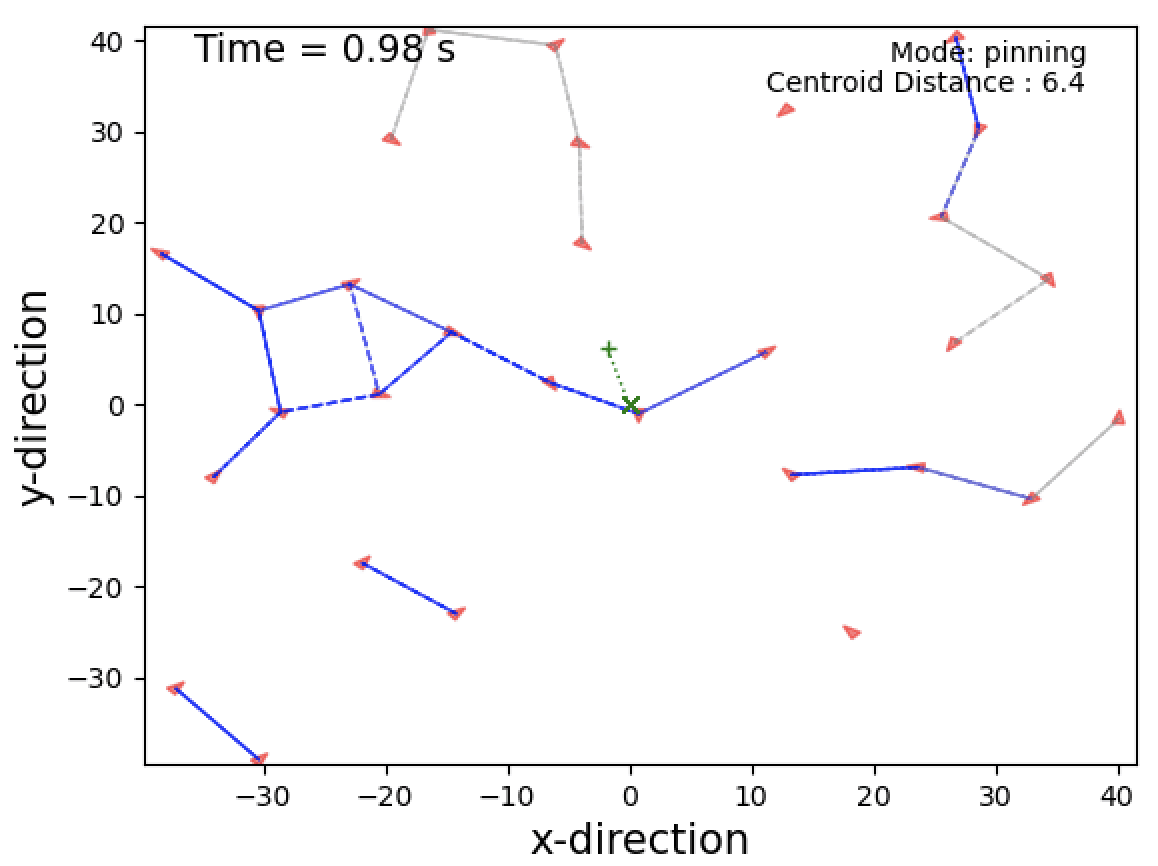}
        \caption{Initial state (disconnected).}
    \end{subfigure}
    \begin{subfigure}[t]{0.45\columnwidth}
        \centering
        \includegraphics[width=\textwidth]{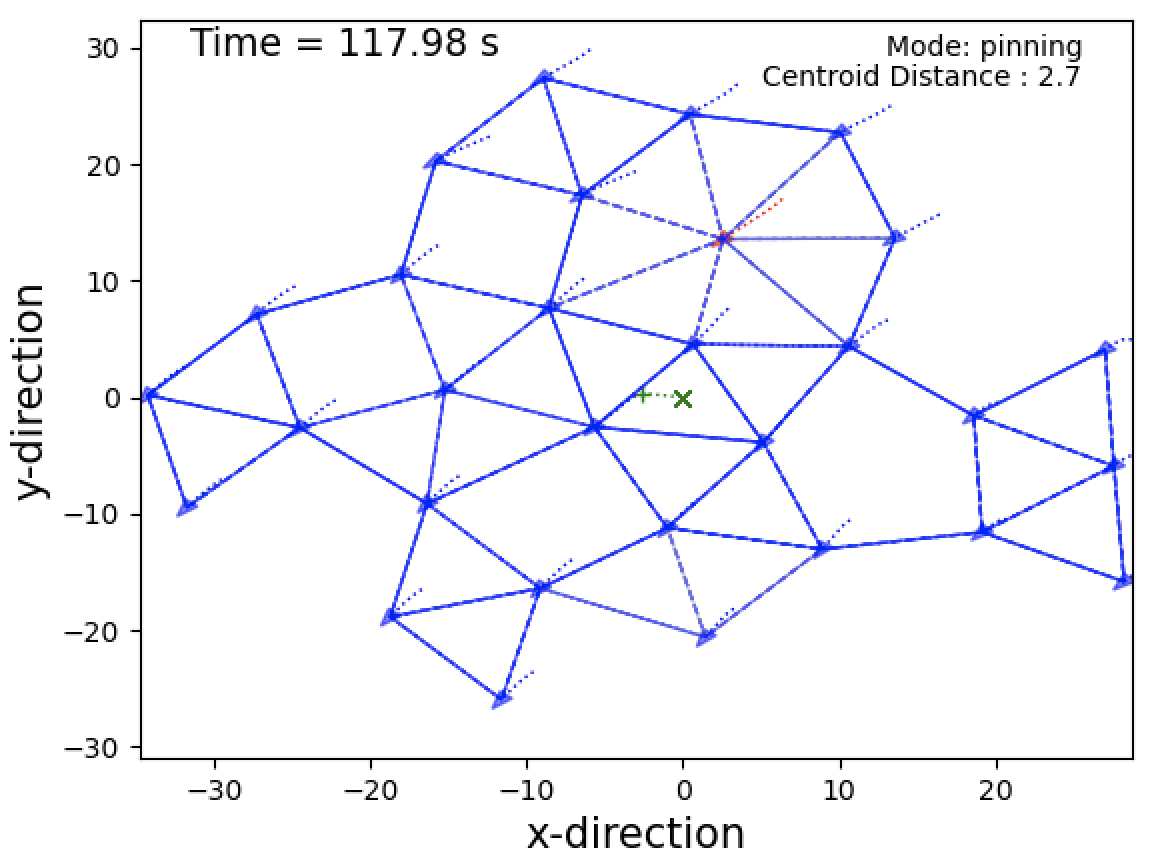}
        \caption{Stabilizes at equilibrium.}
    \end{subfigure}
    \caption{Assembly of 30 agents into flock with $d_{ij}=10$.}
    \label{fig:no_break}
\end{figure}

Homogeneous lattice flocking as described in Section \ref{sec:lattice} was implemented for a fixed value of $d_{ij}=10$. The parameters used for this simulation are summarized in Table \ref{tab:table1}. A total of 30 agents were initiated randomly as shown in Fig.~\ref{fig:no_break}(a). After approximately 90 seconds, the flock stabilized around the positions shown in Fig. \ref{fig:no_break}(b).
\begin{table}[!ht]
    \centering
    \caption{Parameters for Homogeneous Simulation}
    \begin{tabular}{c|c}
        \hline
        Parameter & Value(s)\\ \hline \hline
        Number of agents & $30$ \\
        Flocking parameters & $k_c=10^5$ \\ 
        Desired separation & $d_{ij} = 10$ \\ 
        Maximum sensor range & $r=13$ \\ 
        Sample time & $t_s = 0.02~s$ \\
        Navigation gains & $k_{n,x} = 2.0$, $k_{n,v} = 4.5$ \\
        \hline
    \end{tabular}
    \label{tab:table1}
\end{table}

\begin{figure}[!ht] 
    \centering
    \includegraphics[width=0.90\columnwidth]{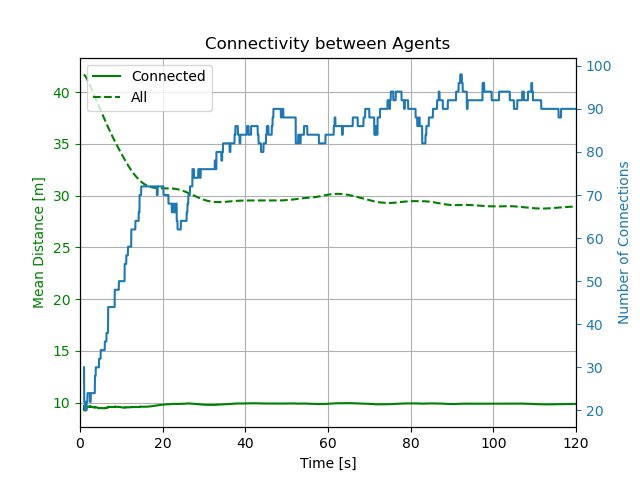} 
    \caption{Agents converge and improve connectivity over time.}
    \label{fig:sim1_connect}
\end{figure}

Fig.~\ref{fig:sim1_connect} illustrates the growth in overall connectivity over time (in blue) and the mean distance between connected agents converging to the desired value of 10 (in solid green). Also note that the overall mean distance between all agents in the flock reduces as the graph components merge into a single component. These results also serve as an illustration Assumption \ref{assum:Vtilde1}.

\subsection{Heterogeneous Assembly}

\begin{figure}[!ht]
    \centering
    \begin{subfigure}[t]{0.45\columnwidth}
        \centering
        \includegraphics[width=\textwidth]{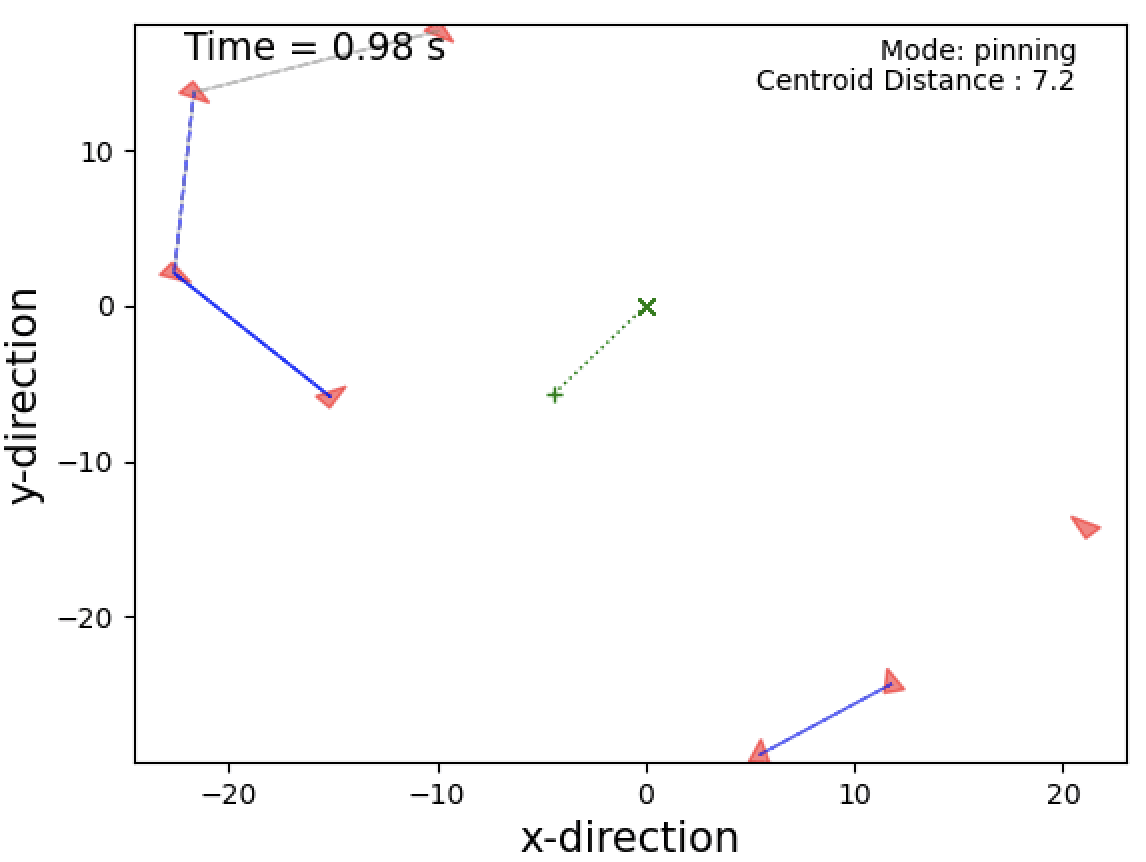}
        \caption{Initial state conflicting separation values (disconnected).}
    \end{subfigure}
    \begin{subfigure}[t]{0.45\columnwidth}
        \centering
        \includegraphics[width=\textwidth]{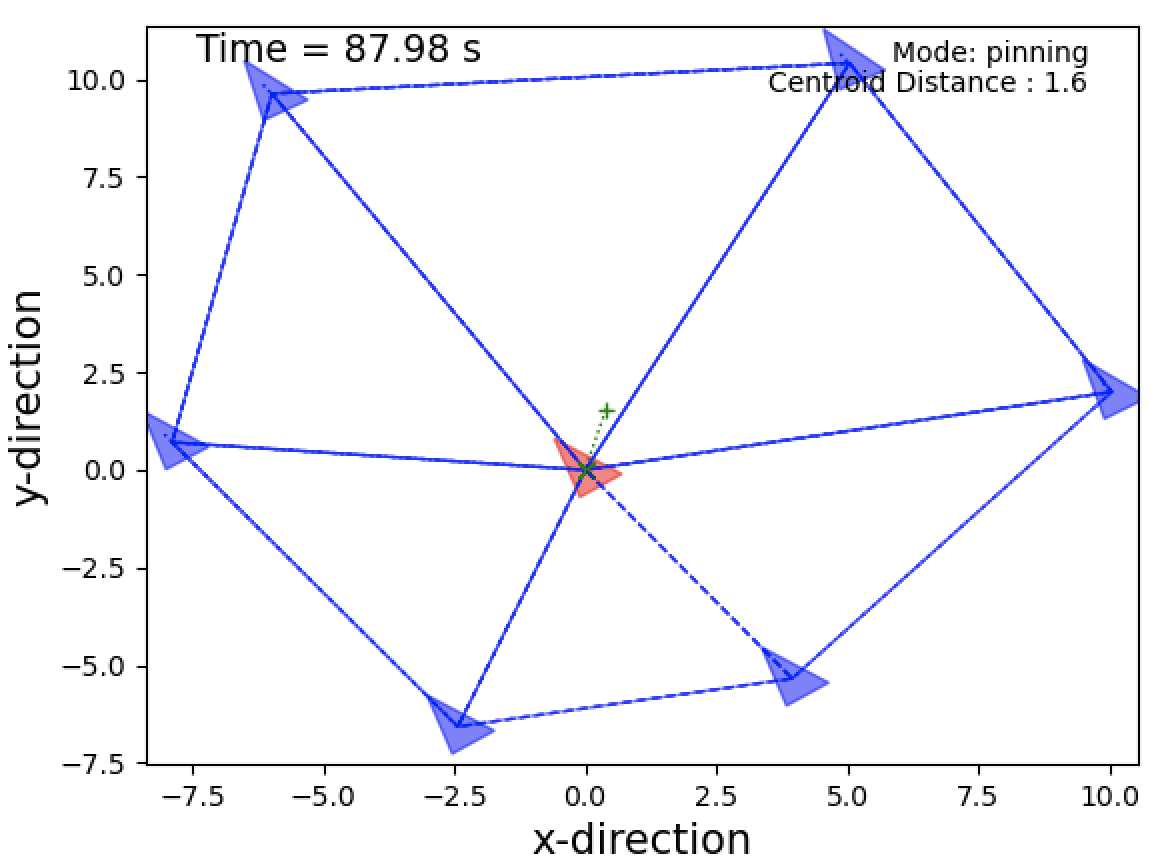}
        \caption{Stabilizes with negotiated separation values (connected).}
    \end{subfigure}
    \caption{Agents reach equilibrium when configured with different separation distances ($d_{ij}$).}
    \label{fig:no_break2}
\end{figure}

Heterogeneous lattice flocking as described in Section \ref{sec:hetlatscale} was implemented for agents with diverse separation values. The parameters used for these simulations are shown in Table \ref{tab:example2}. A total of $7$ agents were initiated randomly as shown in Fig.~\ref{fig:no_break2}(a), with conflicting values of $d_{ij}(0)$ and different separation constraints ($d_{i,min}$ and $d_{i,max}$). After approximately $60~s$, the flock stabilized as shown in Fig.~\ref{fig:no_break2}(b), with each agent having negotiated a shared $d_{ij}$ and $d_{ji}$ with its neighbours based purely on local observations.
\begin{table}[!ht]
    \centering
    \caption{Parameters for Heterogeneous Simulations}
    \begin{tabular}{c|c}
        \hline
        Parameter & Value(s)\\ \hline \hline
        Number of agents &  $7$ \\
        Flocking parameters & $k_c=10^5$ \\ 
        Initial desired separation & $5 < d_{ij}(0) < 13$ \\ 
        Separation constraints & $d_{i,min} = 6 \pm 1$, $d_{i,max} = 11 \pm 2$ \\
        Maximum sensor range & $r=13$ \\ 
        Filter parameter &  $\tau = 0.5$\\ 
        Bump parameter & $p = 4$ \\
        Lattice gain &  $k_d = 0.2$ \\
        Sample time &  $t_s = 0.02~s$ \\
        Navigation gains & $k_{n,x} = 2.0$, $k_{n,v} = 4.5$ \\        
        \hline
    \end{tabular}
    \label{tab:example2}
\end{table}

\begin{figure}[!ht] 
    \centering
    \includegraphics[width=0.90\columnwidth]{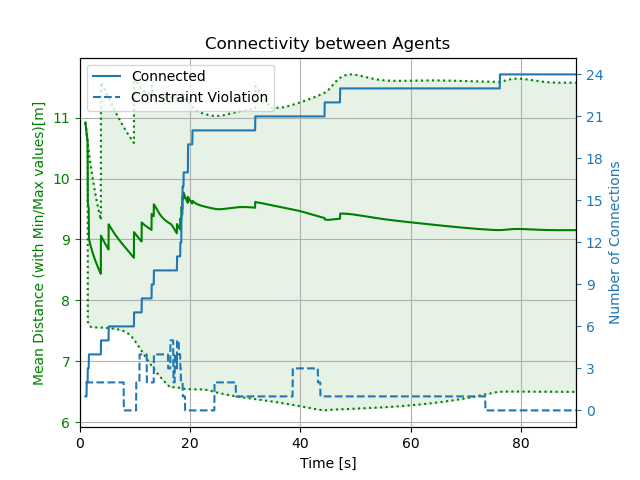} 
    \caption{Agents converge and improve connectivity over time.}
    \label{fig:disconnect2}
\end{figure}

Fig.~\ref{fig:disconnect2} shows the growth in overall connectivity over time (in solid blue). The number of connections involving a potential constraint violation (i.e., a neighbour ($j$) of agent ($i$) has $d_{ji}<d_{i,min}$ or $d_{ji}>d_{i,max}$) is shown in dashed blue. This reduction in potential constraint violations demonstrates the effectiveness of the custom bump function at \eqref{eq:bump} in enforcing constraints. In such cases, agent $i$ is able to model the neighbouring agent's separation parameter using \eqref{filter}, detect the violation of its own constraints ($d_{i,min}$ or $d_{i,max}$), and ignore it in the control policy \eqref{eq:d_c}. The mean distance between connected agents is shown in solid green, with min and max values (reflecting the variance) shown in the shaded green areas. Note the significant variance in parameters, which reach equilibrium within the acceptable bounds. 

\begin{figure}
    \centering
    \includegraphics[width=0.9\columnwidth]{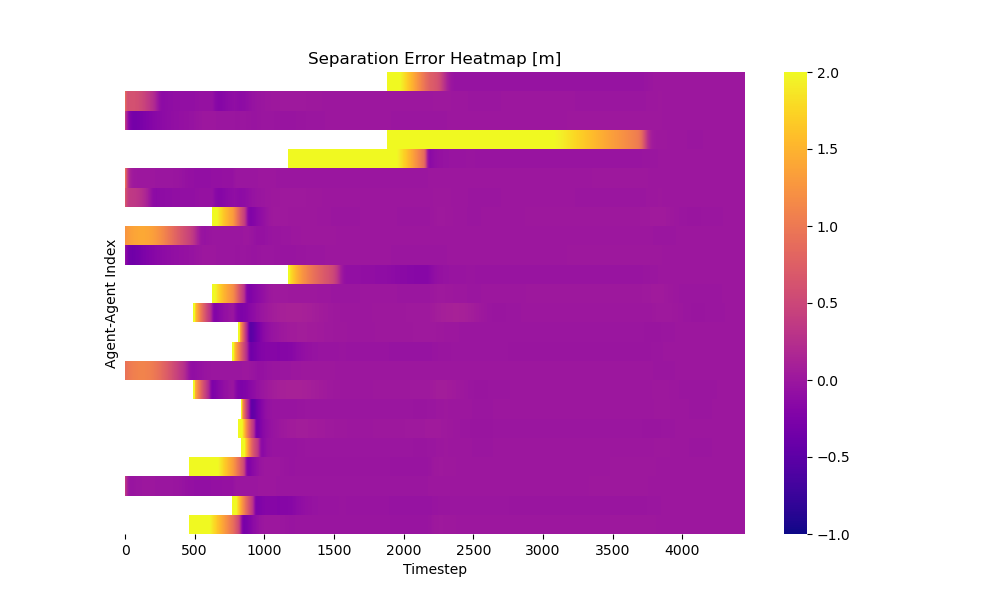} 
    \caption{Heatmap of separation error shows convergence over time.}
    \label{fig:heatmap}
\end{figure}

In order to further illustrate the effectiveness of the approach, Fig. \ref{fig:heatmap} presents a separation error heatmap for another run with the same parameters, but with different initial conditions. The separation error is computed as
$
    e_{s,ij}(t) = \hat{d}_{ji}(t) - d_{ij}(t)~\forall~i\neq j \in \mathcal{V}.
$
 
 Time is shown on the $x$-axis and the index for each unique agent-to-agent connection (i.e., $ij$) is presented as along the $y$-axis. The specific values of $i$ and $j$ have been removed, as they are randomly generated and do not contribute meaningfully to analysis. Many agents start out of range (shown in white). As the flock merges (guided by pinning control), new neighbours come into range with conflicting separation parameters (shown in yellow). These parameters are eventually negotiated, driving the values of $e_s,ij(t)$ to $0$ (shown in magenta). Of particular interest are the transition areas, where we see the broader spectrum of colors reflecting changing values of $d_{ij}(t)$. These transitions are not necessarily monotonically decreasing, due to the unpredictable interactions of neighbouring agents coming into range.

\section{Conclusion}

A broad range of classical flocking methodologies have provided stable, emergent lattice formations. A prevailing assumption of this previous work has been uniform values of inter-agent distances. Many real‐world applications demand greater flexibility, as multi-agent systems become ubiquitous in modern society and the diversity of agent types and configurations grows. Moreover, some applications require agents to operate with a mixture of shared and distinct goals, often without guarantees of trust. Motivated by these challenges, our work revisits flocking by relaxing this assumption of shared inter-agent distances and individual constraints. Building on well‐established formulations for cohesion, alignment, and navigation, we update the existing framework to include a solution that permits negotiation of these parameters. Central to our approach is the construct of a constrained potential function, which negotiates inter-agent parameters. In the spirit of the broader flocking control canon, this negotiation is achieved locally through observations of agent positions and does not require any inter-agent communication or global knowledge. We validate the effectiveness of the approach through a series of simulations demonstrating convergence of flocks with heterogeneous parameters.

\balance
\bibliographystyle{IEEEtran}
\bibliography{Travis}

\end{document}